\pdfoutput=1

\documentclass[11pt]{article}

\usepackage[]{acl}
\usepackage{booktabs}
\usepackage{enumitem}
\usepackage{graphicx}
\usepackage{multirow}
\usepackage{subfigure} 

\usepackage{times}
\usepackage{latexsym}

\usepackage[T1]{fontenc}

\usepackage[utf8]{inputenc}
 
\usepackage{microtype}

\usepackage{inconsolata}
\usepackage{tabularray}

\usepackage{adjustbox}

\usepackage[textsize=scriptsize]{todonotes}

\title{Stanceosaurus 2.0: Classifying Stance Towards Russian and \\ Spanish Misinformation}

\author{Anton Lavrouk, Ian Ligon, Tarek Naous, Jonathan Zheng, Alan Ritter, Wei Xu \\
  College of Computing \\
  Georgia Institute of Technology \\
  \small{
 \texttt{\{antonlavrouk, iligon3, tareknaous, jonathanqzheng\}@gatech.edu; \{alan.ritter, wei.xu\}@cc.gatech.edu}}}

\begin{document}
\maketitle
\begin{abstract}
The Stanceosaurus corpus \citep{zheng2022stanceosaurus} was designed to provide high-quality, annotated, 5-way stance data extracted from Twitter, suitable for analyzing cross-cultural and cross-lingual misinformation. In the Stanceosaurus 2.0 iteration, we extend this framework to encompass Russian and Spanish. The former is of current significance due to prevalent misinformation amid escalating tensions with the West and the violent incursion into Ukraine. The latter, meanwhile, represents an enormous community that has been largely overlooked on major social media platforms.  By incorporating an additional 3,874 Spanish and Russian tweets over 41 misinformation claims, our objective is to support research focused on these issues.  To demonstrate the value of this data, we employed zero-shot cross-lingual transfer on multilingual BERT, yielding results on par with the initial Stanceosaurus study with a macro F1 score of 43 for both languages. This underlines the viability of stance classification as an effective tool for identifying multicultural misinformation.
\end{abstract}

\section{Introduction}
\label{sec:intro}
Misinformation on social media is a highly multicultural phenomenon  \citep{roozenbeek2020susceptibility}. In the ongoing Russia-Ukraine conflict, Russian-language misinformation and propaganda are important weapons used by both sides to influence the opinions of Internet users across the globe. Meanwhile, Spanish-language misinformation is surging unchecked through virtually every online community.\footnote{\href{https://www.theguardian.com/media/2022/oct/06/disinformation-in-spanish-facebook-twitter-youtube}{The Guardian}}
With these issues in mind, we seek to create a dataset that can help identify Spanish and Russian misinformation beyond a binary yes/no approach. We do this by expanding the Stanceosaurus dataset \citep{zheng2022stanceosaurus} to include Spanish and Russian tweets annotated using a 5-way stance labeling schema  (\citealt{gorrell2018rumoureval}, \citealt{Schiller2021}), thus creating \textit{Stanceosaurus 2.0}.  By fine-tuning multilingual BERT \citep{devlin2019bert}, we experiment with zero-shot cross-lingual transfer, demonstrating the potential for \textit{Stanceosaurus 2.0} to help drive forward misinformation research on Spanish and Russian. Furthermore, recent Twitter policies have made it clear that the site is moving away from account-based labeling of misinformation.\footnote{\href{https://help.twitter.com/en/resources/addressing-misleading-info}{Twitter}} Our dataset presents the opportunity to identify potential misinformation on a per-tweet basis, allowing users to see relevant context for potentially misleading tweets. Some may argue that in recent times, Twitter (now X at the time of revision) has taken a far more "hands-off" approach to misinformation. While this may or may not be true, this dataset can be used on social media platforms that are different from Twitter/X. One can get around the tweet length limit by simply concatenating various tweets, etc. In the following sections, we discuss  what exactly Russian and Spanish misinformation entail and why they are so important. 

\begin{figure}[t]
    \centering
    \includegraphics[width=0.98\linewidth]{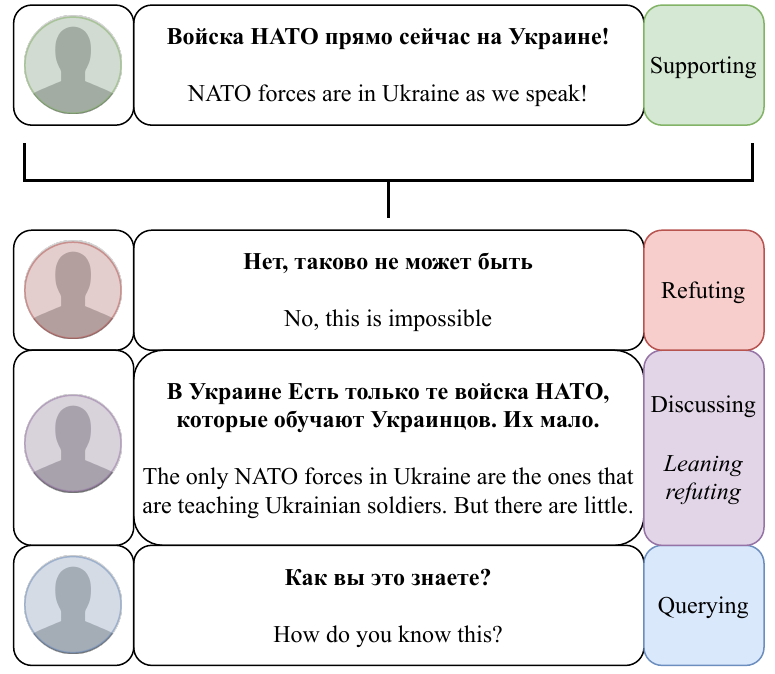}
    \caption{Example of a data point (tweet and context) in the Russian Stanceosaurus dataset. For the claim "NATO forces are currently fighting in Ukraine", we have an example tweet chain demonstrating various stances.}
    \vspace{-.5cm}
    \label{fig:example}
\end{figure}


\paragraph{Russian Misinformation} Misinformation and propaganda are crucial to Russian political warfare. Part of so-called ``active measures'', they are designed to ``weaken the West [and] to drive wedges in the Western community alliances of all sorts, particularly NATO ...'' \citep{alexander2017disinformation}. When Russia launched a full-scale invasion of Ukraine in February of 2022,\footnote{\href{https://www.cnbc.com/2022/02/24/russian-forces-invade-ukraine.html}{CNBC}} both sides of the conflict engaged in hybrid warfare, putting an equal focus on the information front and global deception.\footnote{\href{https://www.theatlantic.com/technology/archive/2022/03/russia-ukraine-war-propaganda/626975/}{The Atlantic}} With propaganda machines in full force, the war in Ukraine has spawned many new misinformation claims. In this context, although a Russian stance dataset is present in \citet{https://doi.org/10.13140/rg.2.2.15252.76161/1}, it is limited, and our research aims to modernize Russian stance data to include wartime misinformation. This is because Russian misinformation "then" and "now" are two different beasts. Potential feasibility for the idea of Russian Stanceosaurus can be seen via the findings in \citet{park2022challenges} which, among many interesting things, identified Twitter as a platform with a significant amount of Russian-language discussion regarding wartime events. The series of \citet{Solopova2023} and \citet{solopova-etal-2023-evolution}, which showcases great results on pro-Kremlin propaganda detection, also potentially implies feasibility of our stance-based approach.

\paragraph{Spanish Misinformation} Misinformation is rampant in the Spanish-speaking world, surging through various online communities \citep{bonnevie2023lessons}. Despite being the fourth most spoken language in the world and an enormous medium for the spread of information worldwide both true and false, misinformation in Spanish is far more of a problem than in English\footnote{\href{https://www.washingtonpost.com/outlook/2021/10/28/misinformation-spanish-facebook-social-media/}{Washington Post}}. This problem is further exacerbated when it is ignored; Facebook whistle-blower Frances Haugen revealed an enormous disconnect between the proportion of users who speak Spanish and the amount of spending committed to anti-misinformation resources in this language\footnote{\href{https://www.theguardian.com/media/2022/oct/06/disinformation-in-spanish-facebook-twitter-youtube}{The Guardian}}. Unsurprisingly, works such as \citet{posadas2019detection} and \citet{fi12050087} attempt to help solve this crucial problem, particularly framing the problem as detecting fake news. These studies show that this kind of claim-based misinformation detection works quite well. Our approach was inspired by such studies. To our knowledge, there are two existing Spanish stance datasets. One is \citet{zotova-etal-2020-multilingual}, a valuable but singular claim-limited collection of Spanish-language stance data. The other is \citet{toledo-ronen-etal-2020-multilingual}, which creates a wonderful Spanish stance dataset, but based on arguments and not misinformation claims. We aim to expand the set of Spanish misinformation via the five-way and three-way classification framework of Stanceosaurus.



\section{Stanceosaurus 2.0: Details}
\label{sec:stance_details}
In order to facilitate the study of Russian and Spanish misinformation, we have created a 5-way stance classification dataset in accordance with the guidelines established by \citet{zheng2022stanceosaurus}. These stance categories are \texttt{Irrelevant}, \texttt{Supporting}, \texttt{Refuting}, \texttt{Querying}, and \texttt{Discussing}. The stance of a tweet is based on the misinformation claim it is discussing. 
An example of various misinformation claim related tweets and their stance categorization can be seen in Figure \ref{fig:example}.  Details on the five stance categories (and how they can be merged to 3 categories) are listed in appendix \ref{sec:appendixC}.


\subsection{Data Collection}

\paragraph{Misinformation Claims} We derived 18 examples of Russian-language misinformation, with 13 from the European Union initiative, \href{https://euvsdisinfo.eu/}{\textit{euvsdisinfo}}, and manually translated them into Russian using a bilingual Russian/English speaker (both fluent). Despite criticism of fact-checking methodology \citep{Giorio2018}, euvsdisinfo is to our knowledge the best source of prominent misinformation which can be found on Russian-language Twitter, especially considering that there is no reliable Russian fact-checking website (this is the reason why we had to translate the claims to Russian). Nonetheless, to mitigate this bias, we supplemented these misinformation claims with 5 claims from the Western media. Again, the absence of claims from Russian sources or Russian-language fact-checking sites is notable. We re-iterate that identifying misinformation is challenging when Russian media is largely state controlled\footnote{\href{https://foreignpolicy.com/2022/04/05/russia-media-independence-putin/}{ForeignPolicy.com}}. Any source that does fact-checking that might disagree with the state media would most likely get taken down or blocked by \href{https://en.wikipedia.org/wiki/Roskomnadzor}{Roskomnadzor}. Therefore, for this study, we selected a ground truth based on western-leaning sources to assess Russian misinformation claims. For the Spanish corpus, we collected 23 misinformation claims from reputable Spanish-language fact-checking websites \href{https://verificado.com.mx/}{\textit{Verificado}}, \href{https://chequeado.com/}{\textit{Chequeado}}, \href{https://www.newtral.es/}{\textit{Newtral}}, and \href{https://chequeabolivia.bo/}{\textit{ChequeaBolivia}}, including claims with various veracity ratings. The selection of both Spanish and Russian claims was guided by the volume of relevant Twitter discourse. The detailed Russian and Spanish claims are listed in Appendix \ref{sec:appendixA} and \ref{sec:appendixB}, respectively. 

\paragraph{Tweet Collection \& Reply Chains} For both languages, we collect tweets using the Twitter API. Queries (Appendix \ref{sec:appendixA} and \ref{sec:appendixB}) are manually curated and iteratively refined in order to capture as many relevant tweets as possible while allowing for diversity in stance categories. This refinement was done by scraping data for a claim using a certain query, and then sampling 20 tweets and hand annotating them. Queries were then modified, and the process was repeated until a reasonably equal distribution of stances was achieved. Alongside tweets from queries, we also collect additional tweets for context, including those ``above''(preceding) and ``below''(following) in the reply chain. Similar to the original Stanceosaurus paper, context tweets were included to potentially help models make classification decisions. Annotation was done by sampling 50 tweets for each claim, adding up to 100 context tweets, which were then evaluated. These quantities were chosen based on the number of available annotators. All three annotators are college-educated students who are fluent in their respective languages. There were two annotators for Russian and one for Spanish. Future versions of this work will include another annotator for Spanish. The 5-class Cohen Kappa for the Russian data is 77.4. Reproducibility criteria for this process are discussed in Appendix \ref{sec:appendixE}. A basic overview of corpus statistics is in Table \ref{tab:small_stats_table}. More detailed corpus statistics  can be found in \ref{sec:appendixG}. As mentioned in \ref{sec:appendixE}, the dataset can be requested directly from the authors. Since the tweet ID's it contains can link tweets back to their authors, it is important that these tweets are used only for academic purposes, especially given that some of them present controversial political opinions.

\begin{table}[ht] 
\setlength{\tabcolsep}{4pt}
\centering
\begin{adjustbox}{width=\linewidth}
\begin{tabular}{l|ccccc|r} 
\toprule
\#tweets & \textbf{Refute} & \textbf{Support} & \textbf{Irrel.} & \textbf{Query} & \textbf{Discuss} & \textbf{Total} \\ \midrule
\textbf{Russian} & 119 & 332 & 999 & 50 & 407 & 1907 \\
\textbf{Spanish} & 270 & 302 & 1036 & 16 & 342 & 1966 \\
\bottomrule
\end{tabular}
\end{adjustbox}
\caption{Corpus statistics  of Stanceosaurus 2.0.}
\label{tab:small_stats_table}
\end{table}


\subsection{Russian Corpus}
\label{sec:russian_corpus}

\paragraph{Russian Twitter} According to a Statista study \citep{statista2023russia}, only 5\% of Russians surveyed reported using Twitter. This makes sense, as before this study, ``Facebook, Instagram, and Twitter were all blocked by the Russian state in early March 2022 when the laws on antiwar activity came into force'' \citep{mccarthy2023four}. Thus, the only way to access Twitter in Russia is through a VPN. From this information, we gather that Russian speakers on Twitter either use a VPN, are native to another country where Russian is a common language, or live abroad. Acknowledging this is important, as it provides a population for the Twitter users that were sampled and presents a limitation of the data to be addressed with future work.

\paragraph{Code Switching} There are instances of tweets that contain different languages. It is fairly common to see acronyms such as ``HIMARS'' and ``NATO'' being written in both English and Russian interchangeably. A brief analysis using regular expressions finds that tweets containing characters from the English alphabet make up about 12 percent of all tweets. Furthermore, sometimes the Russian language is phonetically written in the Latin alphabet. This poses a challenge when querying for relevant Tweets, which we addressed by accounting for as many code-switched variations as was reasonable. Furthermore, in the sampled reply chains, it was common to see the mixing of languages, especially between Russian and Ukrainian. Fortunately, since every single tweet was hand annotated by a fluent Russian speaker with proficient knowledge of Ukrainian, it was not difficult to differentiate between the two languages, as well as any other language that uses the Cyrillic alphabet. 


\paragraph{Obscenities} Due to the politically charged nature of the Russian misinformation claims, many tweets contain large amounts of cursing, which is known as \textit{mat} (pronounced maht). It is argued that \textit{mat} ``is not merely an accumulation of obscenities, but rather constitutes a set of refined, complex structures'', hinting at a ``potentially limitless quantity of expressions'' \citep{Dreizin1982}. A cursory analysis indicates that around 10 percent of all Russian tweets collected contain some sort of obscenity. Context is key when trying to understand Russian obscenities, and this may prove to be quite confusing for a language model to interpret. 

\subsection{Spanish Corpus}
\label{sec:spanish_corpus}

\paragraph{Circumventing Filters} Particularly when discussing the COVID-19 vaccine, many tweets include language that is most likely obscured to circumvent misinformation filters. When searching ``vacuna'' (Spanish for ``vaccine''), Twitter sends users a warning to check federal websites for information related to the pandemic.\footnote{This is no longer the case following \href{https://blog.twitter.com/en_us/topics/company/2021/updates-to-our-work-on-covid-19-vaccine-misinformation}{recent changes to Twitter policy.}} Accordingly, the queries had to be adjusted to include numerous alternative spellings for the word for vaccine (including `vacuno', `vacun@', `vakuna', `cacunados', `v@cunad0s', `kakuna', etc.). 

\paragraph{Social Media Usage} The decision to utilize Twitter for this corpus was driven by its accessible API and publicly shareable text-centric content for open and ethical NLP research. It is worth noting that within the Spanish-speaking realm, Twitter ranks behind Facebook, Instagram, and TikTok in terms of social media usage (\citealt{statista2023}, \citealt{statcounter2023}). Additionally, more Hispanics use WhatsApp than any other race or ethnicity,\footnote{\href{https://www.insiderintelligence.com/content/whatsapp-beats-out-instagram-and-twitter-among-us-hispanic-users}{Insider Intelligence}} and significant volumes of misinformation spread on private channels such as WhatsApp\footnote{\href{https://misinforeview.hks.harvard.edu/article/can-whatsapp-benefit-from-debunked-fact-checked-stories-to-reduce-misinformation/}{Harvard Kennedy School}} where misinformation detection is much more difficult and misinformation is less likely to be corrected by the public.

\paragraph{Code Switching} Mixing Spanish and English together in a single tweet is common, particularly in Spanish-speaking communities in Northern Mexico and the USA. The spread of misinformation in bilingual communities is a unique challenge of particular importance in the United States, where more than one-third of all Hispanic adults self-identify as bilingual in English and Spanish \citep{pewresearch_2015}. 



\section{Automatic Stance Detection Using Stanceosaurus 2.0}

\paragraph{Zero-Shot Cross-Lingual Transfer} In accordance with the original Stanceosaurus paper \citep{zheng2022stanceosaurus}, we conduct a zero-shot cross-lingual transfer experiment on our data. This entails training a model on the English Stanceosaurus dataset of 20,707 tweets and then evaluating it on the Russian and Spanish sets. We believe that this is the best way to evaluate Stanceosaurus 2.0 since we assume that there is little to no stance-based training data available for Russian and Spanish (something we observed during our research, and can be seen in Section~\ref{sec:intro} where we discuss related work). Also, various studies such as \citet{pires2019multilingual} and \citet{Artetxe_2020} have shown zero-shot cross-lingual transfer to be an effective approach in many languages, including Russian and Spanish.

\paragraph{Multilingual BERT} Multilingual BERT \citep{devlin2019bert}, or mBERT, has been shown to be very competitive in the zero-shot setting that we have described (\citealt{wu2019beto}, \citealt{libovický2019languageneutral}). We believe that mBERT is a simple baseline that indicates the quality of our dataset and model performance. For our experiments, we follow the original Stanceosaurus paper \citep{zheng2022stanceosaurus} and use the five stance label schema. To create model input, we format our strings using special tokens as follows: ``[CLS] claim [SEP] text''. 

\paragraph{Loss Functions} Similar to the original Stanceosaurus \citep{zheng2022stanceosaurus}, we examine three different loss functions: cross-entropy loss, weighted cross-entropy loss \citep{Cui_2019_CVPR}, and class-balanced focal loss \citep{baheti-etal-2021-just}. While the cross-entropy loss is a baseline commonly used in classification tasks, we use weighted cross-entropy to modify this baseline to account for imbalanced classes by assigning more weights to classes with fewer samples. Class-balanced focal loss is an alternative method to account for imbalanced classes. It down-weights easy examples and focuses more on difficult ones \citep{Cui_2019_CVPR}.

\paragraph{Results} The results of our experiments can be seen in Table \ref{tab:results}. One can compare these results to English performance on BERT$_{BASE}$ for unseen claims from the original Stanceosaurus paper \citep{zheng2022stanceosaurus}, as well as the same zero-shot cross-lingual transfer experiment on Hindi and Arabic. These extra experiments are also shown in \ref{tab:results}, but they are clearly marked as the contribution of the authors of the original Stanceosaurus paper. Both Russian and Spanish datasets performed similarly to models for English to Hindi and English to Arabic transfer experiments in the original Stanceosaurus \citep{zheng2022stanceosaurus}. The weighted loss functions performed better overall, and both languages achieved an F1 score of around $43$. Reproducibility criteria for our experiments can be seen in appendix \ref{sec:appendixF}.

\begin{table}[h]
\centering
\small
\begin{tabular}{r|ccc}
\multicolumn{4}{c}{\textbf{Russian} (our contribution)} \\
\toprule
\textbf{Loss} & \textbf{Precision} & \textbf{Recall} & \textbf{F1}\\ 
\midrule
CE  & 53.55{\tiny  $\pm$0.8} & 35.33{\tiny  $\pm$0.7} & 36.15{\tiny  $\pm$1.3}\\
Weighted CE & 44.38{\tiny  $\pm$0.2}  & 42.84{\tiny  $\pm$0.5} & 42.09{\tiny  $\pm$0.1} \\
CBFL & 45.60{\tiny  $\pm$1.5} & 46.98{\tiny  $\pm$2.0} & 43.94{\tiny  $\pm$0.2} \\
\bottomrule 
\multicolumn{4}{c}{} \\
\multicolumn{4}{c}{\textbf{Spanish} (our contribution)} \\
\toprule
\textbf{Loss} & \textbf{Precision} & \textbf{Recall} & \textbf{F1}\\ 
\midrule
CE  & 50.26{\tiny  $\pm$1.9} & 40.86{\tiny $\pm$0.7} & 41.81{\tiny $\pm$1.0} \\
Weighted CE & 54.12{\tiny  $\pm$0.4} & 42.65{\tiny  $\pm$0.5} & 43.75{\tiny  $\pm$0.4} \\
CBFL & 51.26{\tiny  $\pm$2.2} & 44.15{\tiny  $\pm$0.9} & 43.83{\tiny  $\pm$1.0} \\
\bottomrule 
\multicolumn{4}{c}{} \\
\multicolumn{4}{c}{\textbf{Hindi} \citep{zheng2022stanceosaurus}} \\
\toprule
\textbf{Loss} & \textbf{Precision} & \textbf{Recall} & \textbf{F1}\\ 
\midrule
CE  & 52.1{\tiny  $\pm$2.9} & 39.4{\tiny $\pm$2.0} & 40.8{\tiny $\pm$2.5} \\
Weighted CE & 55.0{\tiny  $\pm$4.2} & 42.4{\tiny  $\pm$1.4} & 44.3{\tiny  $\pm$1.8} \\
CBFL & 53.0{\tiny  $\pm$3.4} & 44.1{\tiny  $\pm$1.7} & 45.3{\tiny  $\pm$1.5} \\
\bottomrule 
\multicolumn{4}{c}{} \\
\multicolumn{4}{c}{\textbf{Arabic} \citep{zheng2022stanceosaurus}} \\
\toprule
\textbf{Loss} & \textbf{Precision} & \textbf{Recall} & \textbf{F1}\\ 
\midrule
CE  & 44.8{\tiny  $\pm$4.0} & 40.1{\tiny $\pm$2.5} & 40.0{\tiny $\pm$2.0} \\
Weighted CE & 44.1{\tiny  $\pm$3.3} & 40.7{\tiny  $\pm$1.6} & 39.7{\tiny  $\pm$1.7} \\
CBFL & 46.1{\tiny  $\pm$2.6} & 44.7{\tiny  $\pm$1.1} & 43.1{\tiny  $\pm$0.2} \\
\bottomrule 
\multicolumn{4}{c}{} \\
\multicolumn{4}{c}{\textbf{English on BERT$_{BASE}$ } \citep{zheng2022stanceosaurus}} \\
\toprule
\textbf{Loss} & \textbf{Precision} & \textbf{Recall} & \textbf{F1}\\ 
\midrule
CE  & 51.1{\tiny  $\pm$1.1} & 50.5{\tiny  $\pm$2.0} & 50.4{\tiny  $\pm$1.6}\\
Weighted CE & 50.5{\tiny  $\pm$1.9}  & 52.7{\tiny  $\pm$1.1} & 51.3{\tiny  $\pm$1.3} \\
CBFL & 50.6{\tiny  $\pm$1.3} & 55.7{\tiny  $\pm$2.1} & 52.5{\tiny  $\pm$1.0} \\
\bottomrule 
\end{tabular}
\caption{Russian and Spanish experiments. Models are trained on English Stanceosaurus and then evaluated on either Russian or Spanish in our work. F1 is measured as macro F1. Results are taken as the average of 3 experiments, with error being one standard deviation. English, Arabic, and Hindi experiments are taken directly from Stanceosaurus \citep{zheng2022stanceosaurus} as a comparison benchmark.}
\label{tab:results}
\end{table}

\section{Conclusion}

We introduce Stanceosaurus 2.0, an extension of the 5-way stance dataset Stanceosaurus \citep{zheng2022stanceosaurus}. Our dataset includes 18 Russian misinformation claims (1907 tweets) and 23 Spanish misinformation claims (1966 tweets). Our dataset is modern and up to date given the recent slough of misinformation and current events. It also contains Russian and Spanish, which as shown previously, are two languages in which misinformation thrives, and efforts to combat it are limited. Our zero-shot cross-lingual transfer experiments show that our dataset performs at similar levels to that of Hindi and Arabic in the original Stanceosaurus, with a macro F1 score of about 43. This means that there is potential to continue refining models and algorithms to create a somewhat reliable stance classifier using transformer-based models like mBERT. Future versions of this work will entail experiments on more models, as well as a second annotator for the Spanish version.

\section*{Limitations}

\paragraph{The Veracity of Fact-Checked Claims} One of the biggest limitations of our work is the fact that fact-checking is often not as black-and-white as it seems and is generally a practice that suffers from many limitations \citep{doi:10.1080/08913811.2013.843872}. It is very difficult to find objective truths that are verified to a degree of absolute precision for a work like this. This is doubly so for political-leaning claims, such as the claims in the Russian dataset.

\paragraph{Russian Misinformation Claims} An unfortunate limitation of the Russian language is that there are no Russian fact-checking websites that would provide reasonably objective fact-checking, at least as far as we are aware. This is most likely due to the level of control that the Russian government has over the Russian internet \citep{polyakova2019exporting}. This lack of resources means that Russian claims were hand-picked. This could introduce author bias, and may not be an accurate representation of the Russian internet, as claims were mostly all found on the heavily western-leaning website \href{https://euvsdisinfo.eu/}{euvsdisinfo}, as discussed in section \ref{sec:stance_details}.

\paragraph{Russian Twitter} As mentioned in section \ref{sec:russian_corpus}, Twitter is not the most used social media, and this could introduce various biases into our data. Future work could involve the social media website \href{https://vk.com/?lang=en}{VKontakte}, which as mentioned earlier, is the most popular in Russia. However, some problems could arise due to state-owned entities being shareholders\footnote{\href{https://www.reuters.com/article/russia-vk/ceo-of-russias-vk-resigns-as-state-assumes-control-of-internet-firm-idUKL8N2SO3IY}{Reuters}}.

\paragraph{Spanish Twitter} Likewise, Twitter is far from the most popular social media network in Latin America. More work should be done to analyze misinformation on Facebook and WhatsApp in the Hispanosphere. Despite favoring small-group communication, WhatsApp persists as a medium for rapid misinformation dissemination in Latin America \citep{NOBRE2022102757}. 

\paragraph{Spanish Queries} As mentioned in section \ref{sec:spanish_corpus}, numerous obstacles made it difficult to query for relevant Tweets in Spanish. From properties inherent to the Spanish language like a highly inflectional morphology to broader social factors including the prevalence of code-mixing and filter circumvention, care had to be taken when querying Twitter's API to find relevant Tweets without biasing the data in any one direction \citep{1e45ad9f-2b2a-396d-a263-0b0cc52a94d6}. Future work might include broad queries to procure larger datasets that can then be manually cleaned to include more relevant Tweets.

\paragraph{Code Switching} As mentioned in both sections \ref{sec:russian_corpus} and \ref{sec:spanish_corpus}, both languages experienced a decent amount of code-switching, whether it be in the context or the tweet itself. It has been shown before that dealing with code-switching is not an easy task \citep{winata2021multilingual}. However, recently there has been a large number of code-switching datasets that have become available \citep{9074205}. Potential further research may include creating stance datasets exclusively on code-switched datasets.

\paragraph{Tweet Deletion} A feature of the obscured version of the dataset (the version we plan on giving out in most cases) is that it only features tweet IDs. However, if someone deletes a tweet, that tweet will be gone from the obscured dataset. This maintains the user's right to remove their content without it still being a database. However, this may be an issue for researchers using this dataset a long time after the tweets were originally collected. 

\section*{Ethics Statement}
\paragraph{Working With Social Media Data} Mining social media data from Twitter users without their consent is at best ethically problematic \citep{taylor2018mining}. Unfortunately, this kind of data would not exist without this technique. However, our publicly available dataset only contains tweet IDs and does not include actual tweets and usernames. Furthermore, social media data can contain harmful biases towards certain groups, as moderating social media can be extremely difficult \citep{ganesh2020countering}. We encourage a thorough review of the data and its context before deploying in a production environment. 

\paragraph{Data Annotation} We recognize that some of the tweets that have been annotated deal with sensitive topics and contain some hateful language, especially in the Russian dataset, given its political nature. We recognize that annotators need to be warned of this before they start annotating.

\paragraph{Propaganda Analysis} An issue with analyzing propaganda and misinformation is that this analysis can potentially fall into the wrong hands. For example, using this dataset to analyze the effectiveness of Russian propaganda can inform the source of the propaganda exactly what they could improve on. 

\section*{Acknowledgments}
The authors would like to thank Dennis Pozhidaev for his help with data annotation and evaluation. This research is supported by the NSF (IIS-2052498) and IARPA via the HIATUS program (2022-22072200004). The views and conclusions contained herein are those of the authors and should not be interpreted as necessarily representing the official policies, either expressed or implied, of NSF, ODNI, IARPA, or the U.S. Government. The U.S. Government is authorized to reproduce and distribute reprints for governmental purposes notwithstanding any copyright annotation therein.

\bibliography{anthology,custom}

\appendix
\section{Russian Claims and Queries}
\label{sec:appendixA}
Russian claims and queries can be found in figure \ref{fig:RU_claims_queries}.

\section{Spanish Claims and Queries}
\label{sec:appendixB}
Spanish claims and queries can be found in figures \ref{fig:SP_claims_queries_1} and \ref{fig:SP_claims_queries_2}.

\section{Stance Categorization}
\label{sec:appendixC}

The following is a description of each stance:
\begin{itemize}[noitemsep]
    \item \textbf{Supporting:} Tweets that directly support the fact that a claim is true.
    \item \textbf{Refuting:} Tweets that refute the veracity of a claim.
    \item \textbf{Querying:} Questions the veracity of a claim.
    \item \textbf{Discussing:} Provides neutral information on the context or truth of a claim.
    \item \textbf{Irrelevant:} Not relevant to the given claim.
\end{itemize}
If a tweet is labeled as discussing, then to enable 3-way stance classification, the tweet is also given a leaning. The following is a description of each leaning:
\begin{itemize}[noitemsep]
    \item \textbf{Supporting:} The tweet has an indirect positive bias when discussing the claim.
    \item \textbf{Refuting:} The tweet has an indirect negative bias when discussing the claim.
    \item \textbf{Other:} The tweet does not have any sort of bias.
\end{itemize}
With this information, we can now construct our guidelines for 3-way stance categorization as well:
\begin{itemize}[noitemsep]
    \item \textbf{Supporting:} Merge supporting with discussing$_{supporting}$.
    \item \textbf{Refuting:} Merge refuting with discussing$_{refuting}$.
    \item \textbf{Other:} Merge irrelevant, querying, and discussing$_{other}$.
\end{itemize}

\section{Dataset Reproducibility Criteria}
\label{sec:appendixE}
\begin{figure*}[h]
    \centering
    \includegraphics[width=1.05\textwidth]{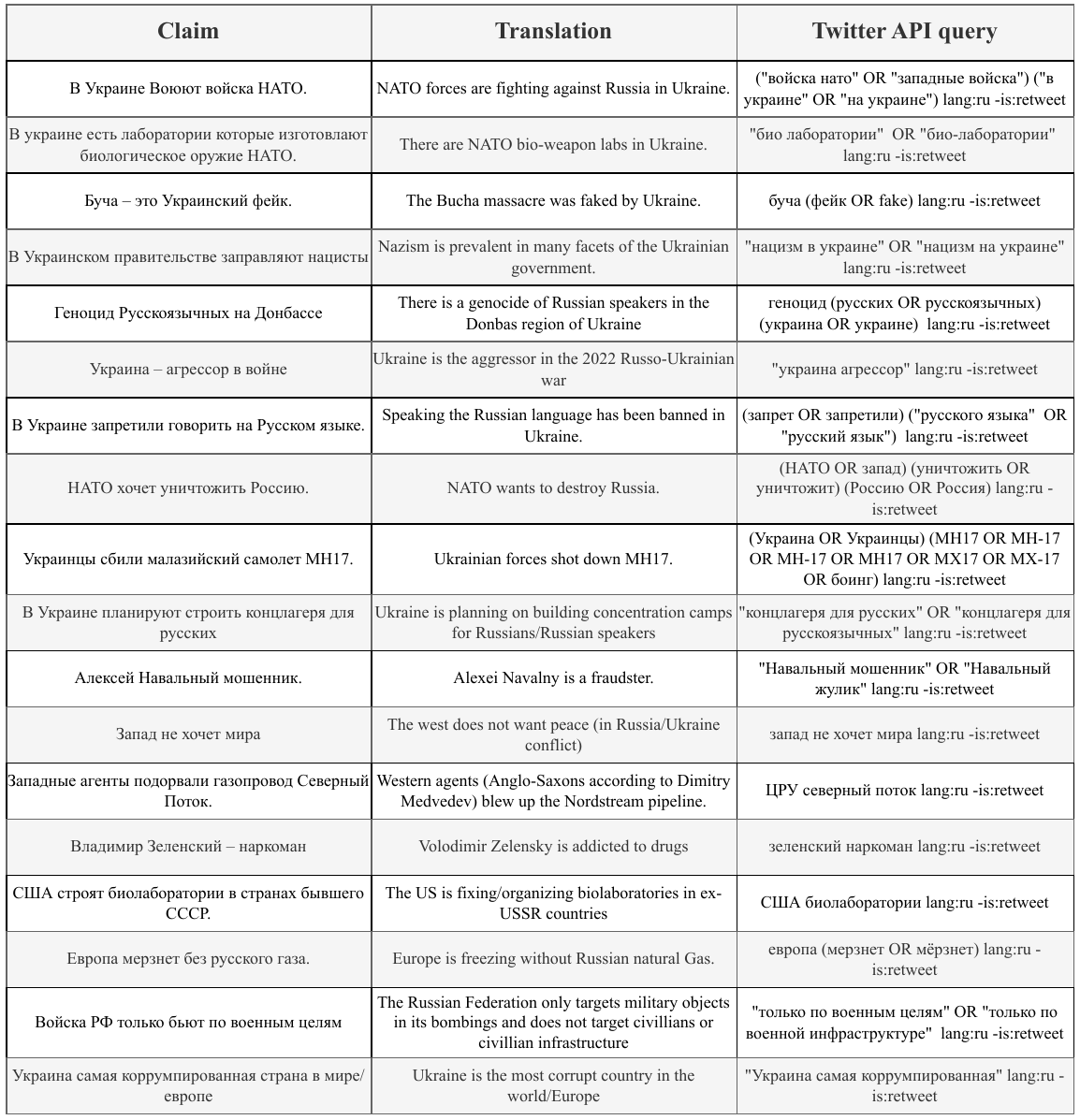}
    \caption{Russian Claims and Queries}
    \label{fig:RU_claims_queries}
\end{figure*}

\begin{figure*}[h]
    \centering
    \includegraphics[width=1.05\textwidth]{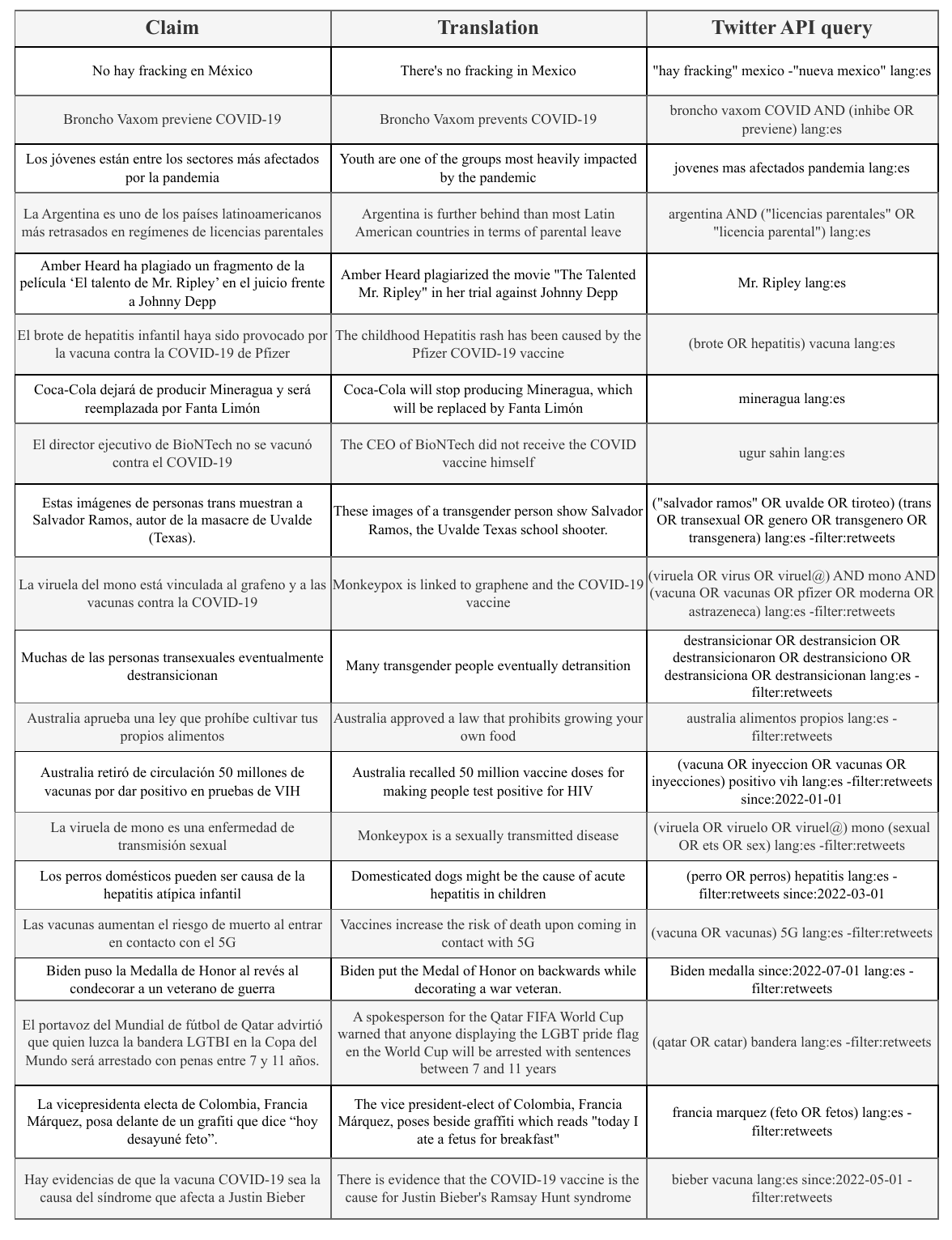}
    \caption{Part 1 of Spanish Claims and Queries}
    \label{fig:SP_claims_queries_1}
\end{figure*}

\begin{figure*}[h]
    \centering
    \includegraphics[width=1.05\textwidth]{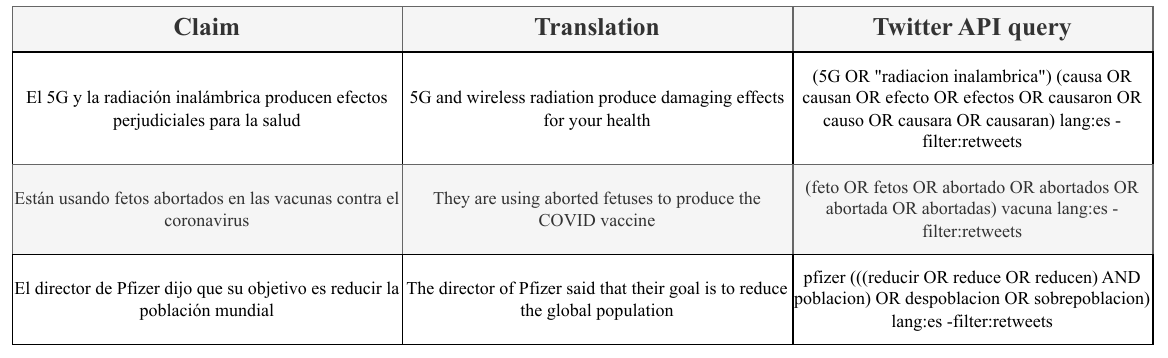}
    \caption{Part 2 of Spanish Claims and Queries}
    \label{fig:SP_claims_queries_2}
\end{figure*}

\begin{itemize}[noitemsep]
    \item Using the twitter API, up to 150 tweets were pulled for each claim using the queries listed in figures \ref{fig:RU_claims_queries}, \ref{fig:SP_claims_queries_1}, and \ref{fig:SP_claims_queries_2}. Context for each tweet was also retrieved. Context in this case means the entire reply chain from the root tweet down to the pulled tweet, as well as any immediate replies.
    \item Quality control was done by an extensive iteration of Twitter API queries. We aimed to make queries such that the distribution of stance categories was reasonably even, although this proved to be difficult with the "Querying" category.
    \item With these tweets in hand, up to 50 tweets were sampled for each claim for annotation. Context tweets were also annotated. Up to 50 parent context tweets were sampled and up to 50 context children tweets were sampled for each claim.
    \item Claims were annotated in accordance with details given in appendix \ref{sec:appendixC}. Russian tweets were double annotated, while Spanish tweets currently only have a single annotator, but we are working to find another annotator at the moment.
    \item Tweets were pre-processed to remove duplicates using lexical similarity.
    \item The context chains were then reconstructed and formatted in json to match the original Stanceosaurus paper \citep{zheng2022stanceosaurus}.
    \item The dataset can be requested from the authors using the emails given in the paper. Since the data is potentially sensitive (tweets of political nature) we need to make sure that anyone who uses these tweets is doing so solely out of academic intent.
\end{itemize}

\section{Experiment Reproducibility Criteria}
\label{sec:appendixF}
\begin{itemize}[noitemsep]
    \item \textbf{Model:} \href{https://huggingface.co/bert-base-multilingual-uncased}{bert-base-multilingual-uncased}
    \item \textbf{Computing Infrastructure:} 4 Nvidia Titan X GPUs. NVIDIA-SMI 460.84. Driver Version 460.84. CUDA version 11.2. Running on CentOS linux 7. Conda version 7. Package versions listed in \texttt{requirements.txt} file in code used.
    \item \textbf{Average Training Time:} Per experiment, around ~40 minutes
    \item \textbf{Evaluation Metrics:} Best evaluation of the development set per training run
    \item \textbf{Number of Experiments:} Each row in \ref{tab:results} was done 3 times. Results are the mean $\pm$ the standard deviation. Random seeds for the three runs were 10, 20, and 30.
    \item \textbf{Hyperparameters:} Hyperparameters were chosen based off of best performing hyper parameters in the original Stanceosaurus model, and then manually tuned.
    \begin{itemize}
        \item \textbf{Learning Rate:} 3e-5
        \item \textbf{Batch Size:} $8$ per GPU, so $32$ total
        \item \textbf{Class Balanced Focal Loss:} Similar to the original paper, we tune $\beta$ and $\gamma$ between $[0.1, 1)$ and $[0.1, 1.1]$ respectively.
        \item The rest are defaulted to what is used in the code. Run commands are included with code.
    \end{itemize}
    \item Code zip file can be accessed upon request.
\end{itemize}

\section{Corpus Statistics}
\label{sec:appendixG}
\begin{figure*}[h]
    \centering
    \subfigure[]{
        \includegraphics[width=0.45\textwidth]{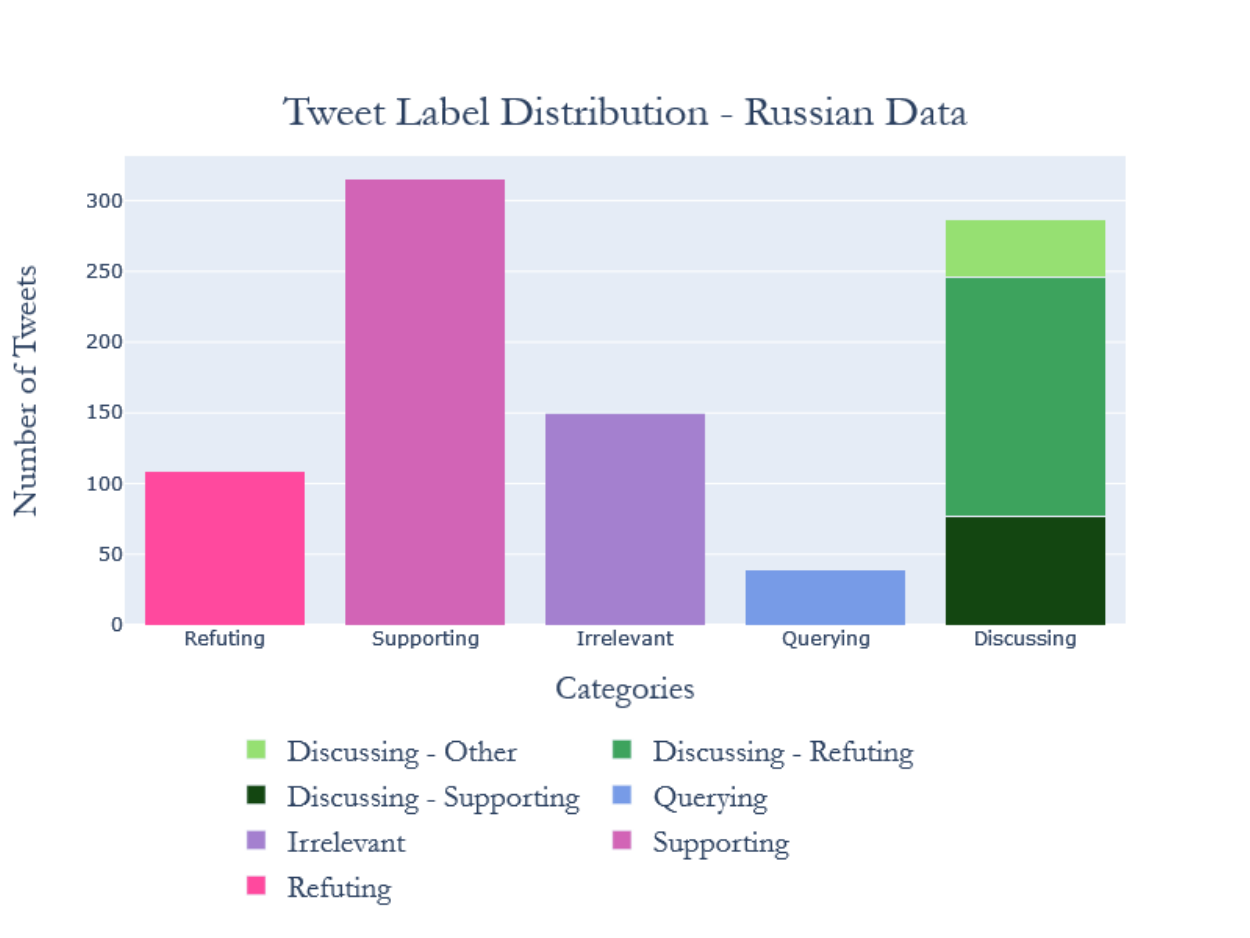}
        \label{fig:RU_bargraph}
    }
    \subfigure[]{
        \includegraphics[width=0.45\textwidth]{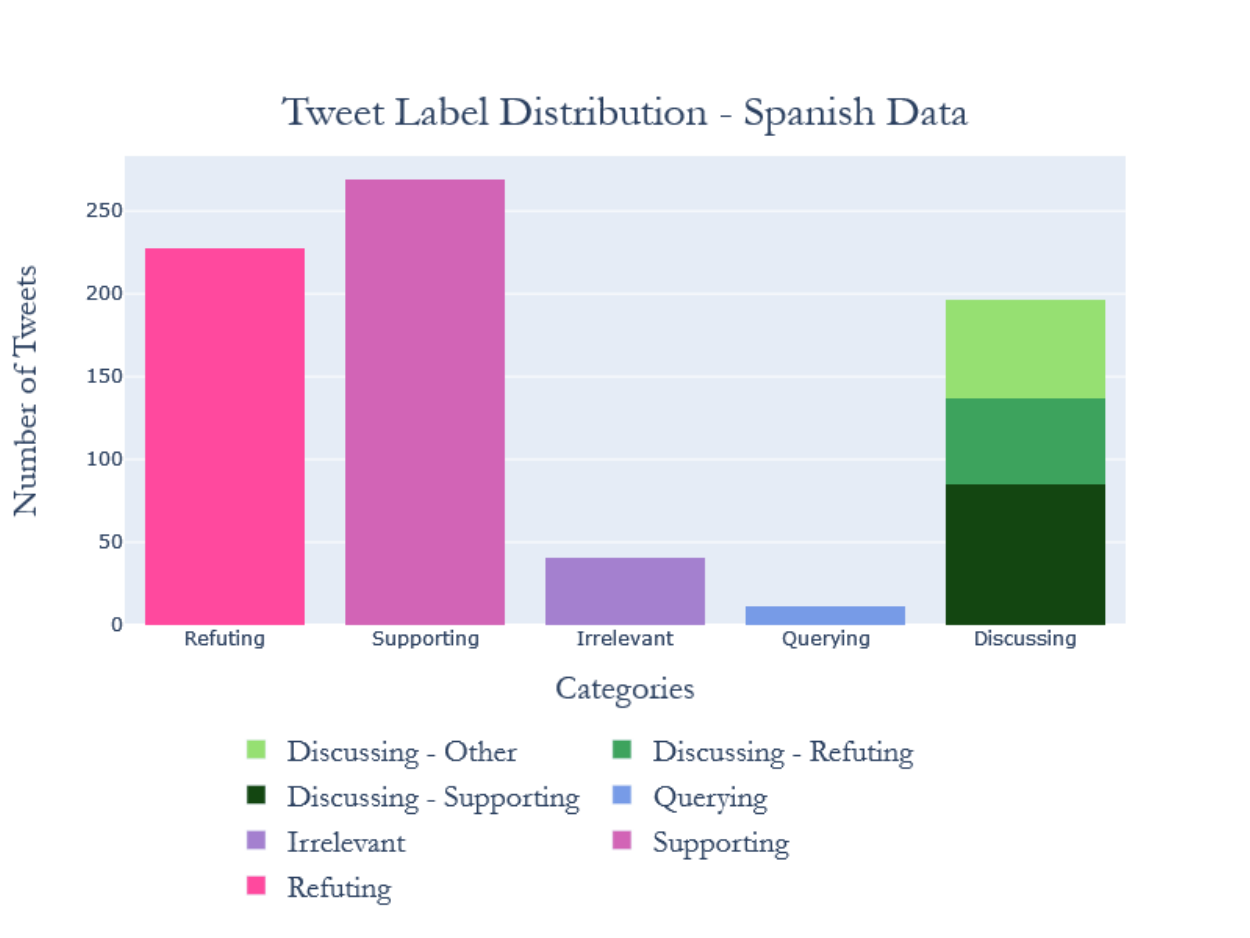}
        \label{fig:SP_bargraph}
    }
    \caption{Label distribution for tweets (by query, not context) in the (a) Russian dataset and (b) Spanish dataset.}
    \label{fig:bargraphs}
\end{figure*}

The distribution of labels and tweet types for Russian Spanish are shown in tables \ref{tab:RU_stats_table} and \ref{tab:SP_stats_table} respectively. A visual representation of the tweets (not context or replies) for Russian Spanish is shown in figures \ref{fig:RU_bargraph} and \ref{fig:SP_bargraph} respectively.

\section{Annotation Logistics}
\label{sec:appendixH}
Annotators were American college students paid 18 dollars an hour. Each annotator was fluent in the language they were annotating. All annotators were recruited as people the authors directly knew. Verbally, annotators were told the scope of the paper and given the abstract.

\section{Use of AI assistants}
\label{sec:appendixI}
AI assistants were used by the authors of this paper in order to proofread the paper. Occasionally, an AI assistant was asked to rephrase some text, just to generate some ideas on sentence flow. Work was never directly copied, and model output was used as inspiration.
\begin{table*}
\centering
\begin{tblr}{
  vline{2} = {-}{},
  hline{2,5} = {-}{},
  vline{9} = {-}{},
}
 & Refuting & Supporting & Irrelevant & Querying & D$_{supporting}$ & D$_{refuting}$ & D$_{other}$ & Total  \\
 Tweets & 109 & 315 & 149 & 39 & 77 & 169 & 41 & 899 \\
 Context & 5 & 15 & 738 & 9 & 51 & 40 & 7 & 865 \\
 Replies & 5 & 2 & 112 & 2 & 6 & 15 & 1 & 143\\
 Total & 119 & 332 & 999  & 50 & 134 & 224 & 49 & 1907 
\end{tblr}
\caption{Russian Corpus Statistics.}
\label{tab:RU_stats_table}
\end{table*}

\begin{table*}
\centering
\begin{tblr}{
  vline{2} = {-}{},
  hline{2,5} = {-}{},
  vline{9} = {-}{},
}
 & Refuting & Supporting & Irrelevant & Querying & D$_{supporting}$ & D$_{refuting}$ & D$_{other}$ & Total  \\
 Tweets & 228 & 269 & 418 & 12 & 85 & 52 & 60 & 1124 \\
 Context & 15 & 21 & 370 & 2 & 18 & 13 & 4 & 443 \\
 Replies & 27 & 12 & 248 & 2 & 76 & 18 & 16 & 399\\
 Total & 270 & 302 & 1036 & 16 & 179 & 83 & 80 & 1966 
\end{tblr}
\caption{Spanish Corpus Statistics.}
\label{tab:SP_stats_table}
\end{table*}



\end{document}